\documentclass[conference]{IEEEtran}
\IEEEoverridecommandlockouts
\usepackage{cite}
\usepackage{amsmath,amssymb,amsfonts}
\usepackage{algorithm}
\usepackage{algorithmic}
\usepackage{graphicx}
\usepackage{textcomp}
\usepackage{xcolor}
\usepackage{multirow} 
\usepackage{url}

\usepackage[numbers]{natbib}

\def\BibTeX{{\rm B\kern-.05em{\sc i\kern-.025em b}\kern-.08em
    T\kern-.1667em\lower.7ex\hbox{E}\kern-.125emX}}
\begin{document}

\title{Reduced Spatial Dependency for More General Video-level Deepfake Detection\\
\thanks{
This work is supported in part by the National Key Research and Development Program of China under Grant 2023YFC3305402, Grant 2023YFC3305401, Grant 2022YFC3303301, the National Natural Science Foundation of China (Nos.62302059 and 62172053), the Shanghai Key Laboratory of Forensic Medicine and Key Laboratory of Forensic Science, Ministry of Justice (KF202420), and the Super Computing Platform of Beijing University of Posts and Telecommunications.\\ * Equal corresponding authors. }

\author{
\textit{Beilin Chu\qquad Xuan Xu\qquad Yufei Zhang\qquad Weike You\textsuperscript{*}\qquad Linna Zhou\textsuperscript{*}} \\\\ School of Cyberspace Security, Beijing University of Posts and Telecommunications, Beijing, China \\ \texttt{\{beilin.chu,sh22xuxuan,zhangyufei,ywk,zhoulinna\}@bupt.edu.cn}
}
}

\maketitle

\begin{abstract}
As one of the prominent AI-generated content, Deepfake has raised significant safety concerns. Although it has been demonstrated that temporal consistency cues offer better generalization capability, existing methods based on CNNs inevitably introduce spatial bias, which hinders the extraction of intrinsic temporal features. To address this issue, we propose a novel method called Spatial Dependency Reduction (SDR), which integrates common temporal consistency features from multiple spatially-perturbed clusters, to reduce the dependency of the model on spatial information. Specifically, we design multiple Spatial Perturbation Branch (SPB) to construct spatially-perturbed feature clusters. Subsequently, we utilize the theory of mutual information and propose a Task-Relevant Feature Integration (TRFI) module to capture temporal features residing in similar latent space from these clusters. Finally, the integrated feature is fed into a temporal transformer to capture long-range dependencies. Extensive benchmarks and ablation studies demonstrate the effectiveness and rationale of our approach.
\end{abstract}


\section{Introduction}
In recent years, the emergence of Deepfake has captured global attention, demonstrating remarkable advancements in the field of deep learning. With the capability to manipulate and create hyper-realistic multimedia content, Deepfake techniques \cite{Deepfakes,thies2016face2face} signify a profound transformation in how humans engage with digital media. However, alongside its potential benefits, Deepfake also raises significant ethical, societal, and security concerns. 

To mitigate threats posed by Deepfake, numerous detection methods have been proposed. Currently, these detection techniques can be broadly categorized into two types: image-level and video-level approaches. Image-level methods typically employ Deep Convolutional Neural Networks (DCNNs) as the backbone to identify subtle artifacts in pixel level \cite{li2020face,liu2020global}. In specific, most of them use strong inductive bias of CNNs towards image styles (i.e. texture), to learn pixel distribution discrepancies between authentic and synthetic images \cite{baker2018deep,hermann2020origins}. As such, numerous experiments exhibit satisfying performances on several public datasets, such as FaceForensics++ , Celeb-DF, and DFDC \cite{rossler2019faceforensics++,li2020celeb,dolhansky2020deepfake}. However, related research has shown that such ability is intrinsically sensitive to unseen domains, since the style of texture may vary among manipulation methods. In a different light, video-level approaches utilize the inconsistency between successive frames, which is caused by ignorance of inter-frame interaction in the manipulation process. Experiments from several works \cite{de2020deepfake,haliassos2021lips} have shown that such inconsistency commonly exists in different types of forgery methods, making it a potentially discriminative clue to generalize across unseen domains. However, recent video-level detectors still suffer from downgrading when tested on unseen domains. Following the observation in \cite{geirhos2018imagenet,hermann2020origins} that CNN tends to introduce texture bias, we argue that the convolution operation in spatial-temporal backbones also inevitably introduces spatial biases, i.e., the network tends to rely more on irrelevant content information, neglecting temporal information. This contradicts the intent of original methodology.

Inspired by this, we aim at learning more general representations of temporal consistency for Deepfake video detection, which is based on a newly designed Spatial Dependency Reduction (SDR) framework. Our framework is multi-branch, utilizing several Spatial Perturbation Branch (SPB) modules composed of Fully Temporal Convolutional Network (FTCN) to extract representations with minimal spatial bias from multiple sets of spatially perturbed samples. Inspired by mutual information theory, we designed the Task-Relevant Feature Integration (TRFI) module, which enhances the network’s ability to extract generalized temporal cues by integrating common temporal consistency features from clusters of perturbed sample features. Additionally, we incorporate contrastive learning to compress the representation space, where spatially perturbed samples distribute, towards the direction of the classification task, aiding TRFI in information integration. Finally, the aggregated temporal feature are fed into a temporal transformer to obtain the final classification label. 
\begin{figure*}[htbp]
\centerline{\includegraphics[height=3.3in]{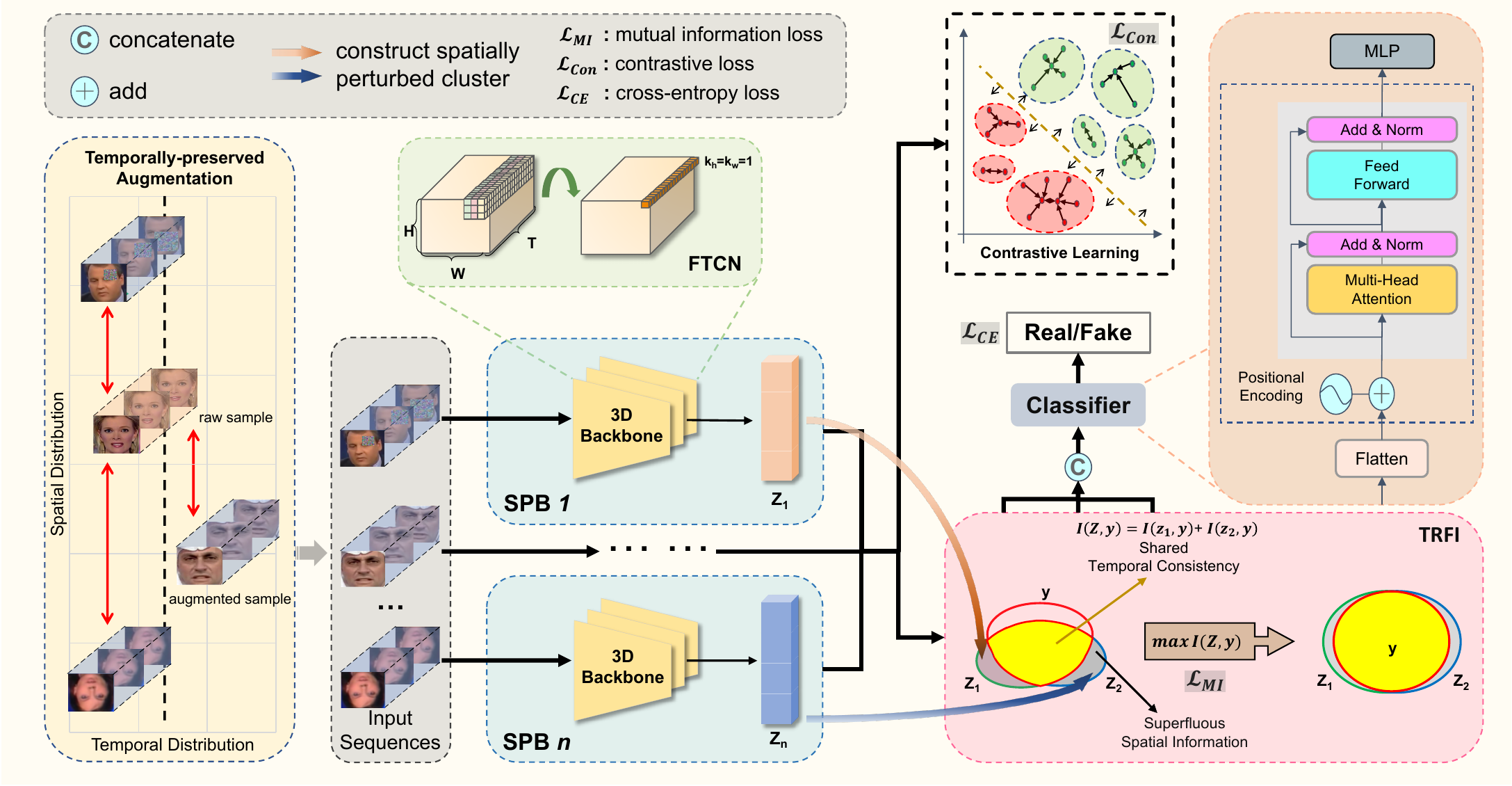}}
\caption{The overall framework of our proposed method Spatial Dependency Reduction (SDR). Samples applied with TPA pass through multi SPBs to construct feature clusters. TRFI diminishes their own spatial distribution while regularizing the shared temporal consistency information.}
\label{fig1}
\end{figure*}

\section{Method}

The proposed method aims to improve the extraction of temporal information by reducing model’s dependency on spatial information. To be specific, we use the strategy of TPA to construct multiple spatially perturbed video sequences and leverage separate FTCNs as backbones to extract multi temporal features. Based on mutual information theory, we propose TRFI to reduce the model’s sensitivity to spatial perturbations while capturing intrinsic and common temporal consistency. Inspired by multi-task learning \cite{liebel2018auxiliary}, we additionally employ contrastive learning with spatially perturbed augmented samples to mitigate noise from latent space, thereby enabling more robust temporal cue extraction. At the final stage, a temporal transformer is introduced to capture long-term dependencies then determine the classification labels. Details of our method are illustrated in Figure~\ref{fig1}.

\subsection{Temporally-preserved Augmentation}
Augmentation is commonly used in Deepfake detection tasks for better generalization \cite{he2020momentum}. In our approach, rather than implicitly enhancing generalization capability as in previous methods, sets of samples applied with distinct augmentation method serve as sufficient feature clusters for subsequent common temporal clue extraction, sharing same temporal features but differing in spatial level. Previous works utilize some common augmentation techniques, such as random clipping, horizontal flipping, and Gaussian noise in image level \cite{chen2021local,sun2022dual}. However, directly incorporating these augmentations into our task would ruin the temporal consistency of original videos. We divide augmentation techniques into two groups. The first type only introduces local spatial randomness and intuitively does not break the motion cues across frames, which could be applied without perturbing global temporal consistency, such as random cutout and color jitter. While the other one visually breaks inter-frame coherence, including random flip and cropping, which needs to be performed cautiously. To this end, for a single video clip, we select one of the following four methods to augment all frames of the sample: Color Jitter, Random Cutout, Flip and Crop. Among them, Flip and Crop are applied with identical angles and ratios, ensuring that the temporal features of the samples remain intact. We refer to this strategy as Temporally-preserved Augmentation (TPA).

\subsection{Spatial Dependency Reduction}
We hypothesise that given a video clip, it has two inherent features—spatial (content information) and temporal (inter-frame consistency), and these two features are partially separable. Our goal is to enable the network to reduce its reliance on spatial features while enhancing the capture of temporal features, which have been shown in related work to provide more generalized cues for detection \cite{de2020deepfake,haliassos2021lips}.

Assume there is a video clip  $c$, and after applying $n$ different TPA techniques, we obtain $\{c_1, c_2, ..., c_n\}$, which are then fed into $n$ separate branches, referred to as the Spatial Perturbation Branch (SPB). We use an existing 3D convolutional backbone as the feature extractor, and following \cite{zheng2021exploring}, we modify the convolutional kernel size from ($K_t \times K_h \times K_w$) to ($K_t \times 1 \times 1$), which further enhances the network’s ability to resist interference from spatial information. After passing through SPB, representations $\{z_1, z_2, ..., z_n\}$ which contain both spatial and temporal information are obtained. Note that, by introducing different spatial perturbation from TPA, in the representation space, $\{z_1, z_2, ..., z_n\}$ have less overlap in spatial representation distribution and more in temporal representation distribution $y$ (as shown in Figure~1 bottom right). Inspired by \cite{ba2024exposing}, we leverage mutual information theory to help the network extract more shared temporal features, which we refer to this way as Task-Relevant Feature Integration (TRFI). Figure~1 (bottom right) shows the information relationship when $n = 2$. In practice, we set $n = 4$, corresponding to the four augmentation methods used in the TPA module.

In the context of mutual information theory, maximizing shared temporal features is given by:
\begin{equation}
    \operatorname* {m a x} I ( y ; \mathcal{Z} ),
\end{equation}
where $I(*)$ is mutual information and $\mathcal{Z}$ represents the joint representation  $\mathcal{Z}=\bigoplus_{i}^{n} z_{i}$. Following the derivation in \cite{ba2024exposing}, this objective is proportional to:
\begin{equation}
    \operatorname* {m a x} \sum_{i=1}^{n} I ( z_{i} ; y \mid\mathcal{Z} \setminus z_{i} ),
\end{equation}
where $\mathcal{Z} \setminus z_{i}$ means representations in set $\mathcal{Z}$ except for the i-th representation. Equation 2 can be approximated using variational methods to avoid the complexity of explicit mutual information calculation:
\begin{equation}
    \sum_{i=1}^{n} I ( z_{i} ; y \mid\mathcal{Z} \setminus z_{i} ) \geq\sum_{i=1}^{n} \operatorname{D_{K L}} \left[ \mathbb{P}_{z} \Vert\mathbb{P}_{z \setminus z_{i}} \right],
\end{equation}
where $P_{Z \setminus z_i} = p(y | Z \setminus z_i)$ and $P_Z = p(y | Z)$ represent the predicted distributions. $\operatorname{D_{K L}}$  denotes the Kullback-Leibler (KL) divergence. The final mutual information loss is computed as:
\begin{equation}
    \mathcal{L}_{\mathrm{M I}}=\operatorname* {m i n}_{\theta} \operatorname{e x p} \left(-\sum_{i=1}^{n} \operatorname{D_{K L}} \left[ \mathbb{P}_{z} \| \mathbb{P}_{z \setminus z_{i}} \right] \right),
\end{equation}
where $\theta$ denotes the model parameters in the TRFI module.

In this way, by integrating temporal feature detection capabilities across different SPBs, our method reduces the model’s reliance on spatial information and enhances the extraction of common temporal cues.

\subsection{Global Temporal Classification}
The Temporal Transformer is designed to capture long-range dependencies along the time dimension and provide the final classification. Following \cite{zheng2021exploring}, we apply a trainable linear projection $\textbf{W}$ to map the feature dimension of integrated representations to the dimension of the transformer, and introduce a trainable [CLS] token as the global representation for final binary classification.

\textbf{Loss Function:}
Inspired by multi-task learning \cite{liebel2018auxiliary}, we perform contrastive learning with samples processed by SPBs to facilitate the TRFI module by reducing latent noise and compressing the latent space towards the direction of binary classification. The contrastive loss function is defined as:
\begin{align}
    \mathcal{L}_{\mathrm{C o n}}=&-\frac{1} {2 | \mathcal{B} |} \sum_{i=1}^{| \mathcal{B} |} \left[ \operatorname{l o g} \frac{\sum_{k \neq i}e^{\mathbf{z}_{r i} \cdot\mathbf{z}_{r k}/ \tau}} {\sum_{k \neq i}e^{\mathbf{z}_{r i} \cdot\mathbf{z}_{r k}/ \tau}+\sum_{j=1}^{| \mathcal{B} |} e^{\mathbf{z}_{r i} \cdot\mathbf{z}_{f j}/ \tau}} \right. \nonumber \\
    & \left. + \operatorname{l o g} \frac{\sum_{k \neq i}e^{\mathbf{z}_{f i} \cdot\mathbf{z}_{f k}/ \tau}} {\sum_{k \neq i}e^{\mathbf{z}_{f i} \cdot\mathbf{z}_{f k}/ \tau}+\sum_{j=1}^{| \mathcal{B}| } e^{\mathbf{z}_{f i} \cdot\mathbf{z}_{r j}/ \tau}} \right] ,
\end{align}
where $z_{ri}, z_{fi}$ denotes real and fake samples respectively and $\tau$ is the temperature hyper-parameter.

Additionally, we employ cross-entropy loss $\mathcal{L}_{\mathrm{C E}}$ after the temporal transformer as the objective function for the final classification head. The total loss to optimize the proposed RSD method is:
\begin{equation}
    \mathcal{L} = \alpha\mathcal{L}_{\mathrm{M I}} + \beta\mathcal{L}_{\mathrm{C o n}} + \gamma\mathcal{L}_{\mathrm{C E}}
\end{equation}
where $\alpha$, $\beta$ and $\gamma$ are the hyper-parameters used to balance corresponding losses.


\begin{table*}[htbp]
\begin{center}
\caption{Video-level generalization tests accuracy (\%) and AUC scores (\%) within FF++.}
\begin{tabular}{c|cccccccccc}
\hline
\multirow{2}{*}{Method} &
  \multicolumn{10}{c}{Training on remaining three} \\ \cline{2-11} 
 &
  \multicolumn{2}{c|}{Deepfake} &
  \multicolumn{2}{c|}{FaceSwap} &
  \multicolumn{2}{c|}{Face2Face} &
  \multicolumn{2}{c|}{NeuralTexture} &
  \multicolumn{2}{c}{Avg} \\ \hline
 &
  AUC &
  \multicolumn{1}{c|}{ACC} &
  AUC &
  \multicolumn{1}{c|}{ACC} &
  AUC &
  \multicolumn{1}{c|}{ACC} &
  AUC &
  \multicolumn{1}{c|}{ACC} &
  AUC &
  ACC \\
Xception \cite{rossler2019faceforensics++} &
  93.7 &
  \multicolumn{1}{c|}{91.3} &
  51.4 &
  \multicolumn{1}{c|}{47.5} &
  87.2 &
  \multicolumn{1}{c|}{86.5} &
  81.0 &
  \multicolumn{1}{c|}{79.4} &
  78.3 &
  76.2 \\
Face x-ray \cite{li2020face} &
  99.5 &
  \multicolumn{1}{c|}{97.4} &
  93.2 &
  \multicolumn{1}{c|}{91.1} &
  94.5 &
  \multicolumn{1}{c|}{93.9} &
  92.5 &
  \multicolumn{1}{c|}{90.8} &
  94.9 &
  93.3 \\
F3-net \cite{qian2020thinking} &
  93.7 &
  \multicolumn{1}{c|}{92.4} &
  94.7 &
  \multicolumn{1}{c|}{93.1} &
  91.8 &
  \multicolumn{1}{c|}{89.3} &
  86.5 &
  \multicolumn{1}{c|}{84.4} &
  91.7 &
  89.8 \\
RECCE \cite{cao2022end} &
  93.5 &
  \multicolumn{1}{c|}{92.6} &
  94.5 &
  \multicolumn{1}{c|}{93.4} &
  91.0 &
  \multicolumn{1}{c|}{90.7} &
  87.4 &
  \multicolumn{1}{c|}{85.6} &
  91.6 &
  90.6 \\
LipForensics \cite{haliassos2021lips} &
  99.7 &
  \multicolumn{1}{c|}{94.9} &
  90.1 &
  \multicolumn{1}{c|}{87.3} &
  99.7 &
  \multicolumn{1}{c|}{98.4} &
  99.1 &
  \multicolumn{1}{c|}{97.6} &
  97.2 &
  94.6 \\
FTCN \cite{zheng2021exploring} &
  \textbf{99.8} &
  \multicolumn{1}{c|}{-} &
  99.6 &
  \multicolumn{1}{c|}{-} &
  99.5 &
  \multicolumn{1}{c|}{-} &
  99.2 &
  \multicolumn{1}{c|}{-} &
  99.5 &
  - \\ \hline
\textbf{SDR (ours)} &
  99.7 &
  \multicolumn{1}{c|}{\textbf{97.5}} &
  \textbf{99.7} &
  \multicolumn{1}{c|}{\textbf{97.9}} &
  \textbf{99.6} &
  \multicolumn{1}{c|}{\textbf{98.6}} &
  \textbf{99.4} &
  \multicolumn{1}{c|}{\textbf{98.7}} &
  \textbf{99.6} &
  \textbf{98.2} \\ \hline
\end{tabular}
\label{tab1}
\end{center}
\end{table*}

\section{Experiments}
\subsection{Experimental Settings}

\textbf{Datasets:} We evaluate our method on the widely-used benchmark dataset FaceForensics++ \cite{rossler2019faceforensics++}, Celeb-DF-v2 \cite{li2020celeb} and DFDC \cite{dolhansky2020deepfake}.

\textbf{Evaluation metrics:} To evaluate the detection capability of our proposed temporal cues exploration, we utilize commonly employed AUC and ACC in video level as metrics.

\textbf{Implementation Details:} We use Retinaface \cite{deng2020retinaface} to detect and crop faces for all the datasets, then resize them to 224 × 224. Each video clip contains 32 frames. The 3D R50 \cite{hara2017learning} is used as backbone in TRFI and the weights of attention heads are randomly initialized. We use a batch size of 16 and Adam optimisation with a learning rate of $10^{-4}$. For more details regarding the FTCN and temporal transformer, we followed the settings in \cite{zheng2021exploring}.

\begin{table}[htbp]
\caption{Video-level generalization tests AUC scores (\%) on the testing datasets after trained on FF++.}
\begin{center}
\begin{tabular}{ccc|c}
\hline
\multirow{2}{*}{Method} & \multirow{2}{*}{Celeb} & \multirow{2}{*}{DFDC} & \multirow{2}{*}{Avg} \\
             &      &      &      \\ \hline
Xception \cite{rossler2019faceforensics++}     & 68.2 & 65.9 & 67.1 \\
Face x-ray \cite{li2020face}   & 75.7 & 67.3 & 71.5 \\
F3-net \cite{qian2020thinking}       & 71.2 & 71.5 & 71.4 \\
RECCE \cite{cao2022end}       & 77.4 & 76.8 & 77.1 \\
SDIF \cite{lai2024selective}        & 81.8 & \textbf{80.7} & 81.2 \\
Local-motion \cite{guan2022delving} & 77.7 & 68.4 & 73.1 \\
Lipforensics \cite{haliassos2021lips} & 82.4 & 73.5 & 78.0 \\
FTCN  \cite{zheng2021exploring}       & 86.9 & 74.0 & 80.5 \\ \hline
\textbf{SDR (ours)}     & \textbf{88.5}          & 76.2         & \textbf{82.4}        \\ \hline
\end{tabular}
\label{tab2}
\end{center}
\end{table}

\begin{table}[htbp]
\caption{Ablation study AUC Scores (\%) on combinations of components.}
\begin{center}
\begin{tabular}{c|c|c|cc}
\hline
TPA                       & TRFI                      & \begin{tabular}[c]{@{}c@{}}Contrastive \\ learning\end{tabular} & Celeb         & DFDC          \\ \hline
                          &                           &  & 86.4 & 72.5 \\
                          & \checkmark &  & 87.9 & 74.3 \\
\checkmark & \checkmark &  & 88.2 & 75.8 \\
\checkmark & \checkmark & \checkmark                                       & \textbf{88.5} & \textbf{76.2} \\ \hline
\end{tabular}
\label{tab3}
\end{center}
\end{table}

\subsection{Comparison with Other Methods}
In this section, we compare the performance with those of the current mature models \cite{qian2020thinking,cao2022end,lai2024selective,guan2022delving}, including video-level ones \cite{haliassos2021lips,zheng2021exploring} and self-supervised method \cite{li2020face}.

We first conduct generalization test on the four forgery types within the FF++ (HQ) dataset. Next, we perform cross-dataset evaluation between datasets with greater sample differences, namely Celeb-DF-v2 and DFDC.

\textbf{Cross-domain Evaluation within FF++.} In this section, we conduct our experiments on four sub-datasets within FF++. Specifically, we train the proposed SDR model with training set of three datasets and then testing the model on the testing set of the remaining one. 

According to Table~\ref{tab1}, RECCE, Lipforensics and FTCN stand out for their generalization ability. However, our method achieves the best result and outperforms FTCN by 0.2\% on average in terms of AUC score. Comparing to Face X-ray and FTCN, we exceed 4.8\%, 4.9\% and 2.5\%, 3.6\%, in AUC and ACC score respectively. In addition, when tested across manipulation methods, many approaches exhibit unstable results, whereas our method maintains stable detection performance. This proves the effectiveness of our method, which ignores spatial biases brought by different forgeries, focusing on intrinsic temporal clues.

\textbf{Cross-domain Evaluation across Datasets.}
Here we conduct our experiments on two datasets Celeb-DF-v2 and DFDC, to further evaluate the generalization ability in a more open scenario, which aligns better with real-world situation. We train the model using FF++ and test on other datasets. 

Analyzing the results presented in Table~\ref{tab2}, our SDR method demonstrates superior performance across both datasets. On Celeb-DF-v2, SDR achieves the highest AUC score of 88.5\%, outperforming the next best method, FTCN, which scores 86.9\%. For DFDC, although SDR performs slightly worse than SDIF, the average performance demonstrates the generalization ability of SDR, which extracts universal temporal consistency representations from FF++, and effectively generalizes to other datasets.
 
\subsection{Ablation Study}
\textbf{Study on different settings}
To verify the effectiveness of our design, we conduct ablation experiments with different component combinations. Notably, removing the TPA module applies augmentations (flip and crop with distinct angles and ratios) to each frame, disrupting inter-frame coherence. If we remove TRFI, TPA will also be removed, and sequences are processed directly with single SPB before being passed to the classifier. According to the results in Table~\ref{tab3}, the complete model achieves the best results. When contrastive learning is removed, performance drops by 0.3\% on CelebDF and 0.4\% on DFDC. This demonstrates that contrastive learning indeed helps reduce latent noise at the bottleneck, enhancing the TRFI’s inductive capability. Next, without the guidance of TAP, the model learns more information unrelated to temporal consistency, resulting in poorer performance. Finally, we remove the crucial TRFI, leading to a significant performance drop of 1.5\% and 1.8\% on CelebDF and DFDC, respectively. This proves that our proposed TRFI effectively induces common temporal consistency features by eliminating redundant spatial information dependence, leading to a more generalizable detection approach.

\begin{figure}[htbp]
\centerline{\includegraphics[width=3.5in]{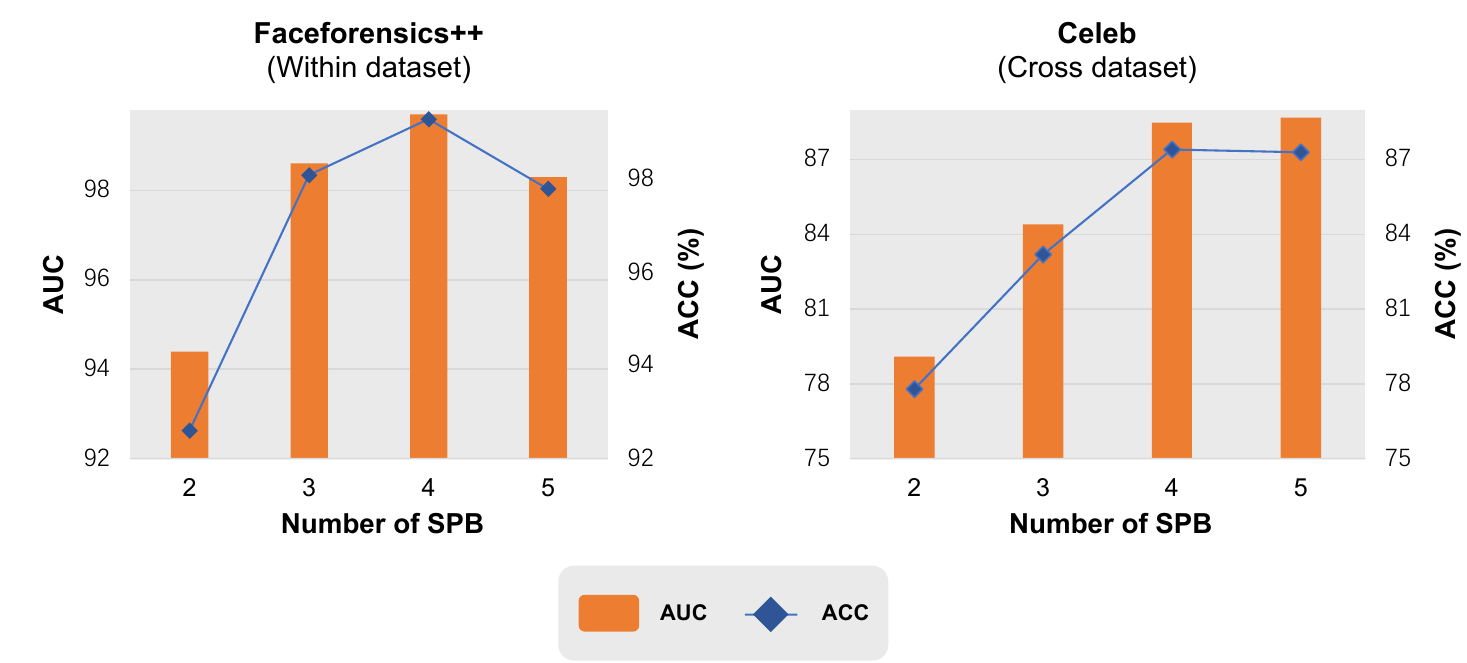}}
\caption{The ablation study on different number of SPBs. For number of 2 and 3, we arbitrarily choose 2 and 3 augmentation methods in TPA. For number of 5, we additionally introduce Gaussian Noise with the same intensity, as an extra SPB.}
\label{fig2}
\end{figure}

\textbf{Study on different number of SPB.}
Figure~\ref{fig2} presents the performance metrics (AUC and ACC) for two datasets, FF++ and Celeb-DF-v2, under different numbers of SPB. 

As depicted in Figure~\ref{fig2}, the number of 4 achieves the best result on both dataset. This proves the branch number of 4 for SPB is a relatively reasonable setting in our method. 

\section{Conclusion}
In this research, we delve into addressing the common temporal consistency exploration in manipulated videos. Foremost, we construct reasonable temporal perturbation samples. Subsequently, we propose a novel method that captures common temporal consistency information through aggregating clusters with identical temporal feature distributions but distinct spatial ones, which could guide spatiotemporal networks to extract more generalized temporal cues. Furthermore, comprehensive experiments underscore the efficacy and rationality of this design in capturing discrepancies in temporal consistency between authentic and fabricated videos.


\newpage

\bibliographystyle{IEEEtran}
\bibliography{IEEEabrv,IEEEexample}

\end{document}